\documentclass[twocolumn, switch]{article} 

\usepackage{preprint}

\usepackage{amsmath, amsthm, amssymb, amsfonts}
\usepackage{bm}
\usepackage{mathtools}

\usepackage[numbers,square]{natbib}
\usepackage[utf8]{inputenc}	
\usepackage[T1]{fontenc}	
\usepackage{xcolor}		
\usepackage[colorlinks = true,
            linkcolor = blue,
            urlcolor  = blue,
            citecolor = blue,
            anchorcolor = black]{hyperref}	
\usepackage{booktabs} 		
\usepackage{nicefrac}		
\usepackage{microtype}		
\usepackage{lineno}		
\usepackage{float}			

\usepackage{newfloat}
\DeclareFloatingEnvironment[name={Supplementary Figure}]{suppfigure}
\usepackage{sidecap}
\sidecaptionvpos{figure}{c}

\usepackage{titlesec}
\titlespacing\section{0pt}{12pt plus 3pt minus 3pt}{1pt plus 1pt minus 1pt}
\titlespacing\subsection{0pt}{10pt plus 3pt minus 3pt}{1pt plus 1pt minus 1pt}
\titlespacing\subsubsection{0pt}{8pt plus 3pt minus 3pt}{1pt plus 1pt minus 1pt}

\usepackage{graphics}
\usepackage{hyperref}
\usepackage{color}
\usepackage{multirow}
\usepackage{hhline}
\usepackage[flushleft]{threeparttable}
\usepackage{times}
\usepackage{epsfig}
\usepackage{graphicx}
\usepackage{amsmath}
\usepackage{amssymb}

\usepackage{subfig}
\usepackage{color}
\usepackage{multirow}

\usepackage{graphicx}
\usepackage{amsmath}
\usepackage{amssymb}

\newcommand{\etal}{\textit{et al}.}

\title{Dual-Modality Vehicle Anomaly Detection via Bilateral Trajectory Tracing}

\usepackage{authblk}

\date{} 

\author[1, *]{Jingyuan Chen}
\author[1, *]{Guanchen Ding}
\author[1, *]{Yuchen Yang}
\author[2]{Wenwei Han}
\author[3]{Kangmin Xu}
\author[1]{Tianyi Gao}
\author[1]{Zhe Zhang}
\author[1]{Wanping Ouyang}
\author[2]{Hao Cai}
\author[1, 2, \dag]{Zhenzhong Chen}
\affil[1]{School of Remote Sensing and Information Engineering, Wuhan University}
\affil[2]{School of Computer Science, Wuhan University, China}
\affil[3]{Hongyi Honor College, Wuhan University, China}

\begin{document}

\twocolumn[ 
  \begin{@twocolumnfalse} 
  
\maketitle

\begin{abstract}
  Traffic anomaly detection has played a crucial role in Intelligent Transportation System (ITS). 
  The main challenges of this task lie in the highly diversified anomaly scenes and variational lighting conditions. 
  Although much work has managed to identify the anomaly in homogenous weather and scene, few resolved to cope with complex ones. 
  In this paper, we proposed a dual-modality modularized methodology for the robust detection of abnormal vehicles.
  We introduced an integrated anomaly detection framework comprising the following modules: background modeling, vehicle tracking with detection, mask construction, Region of Interest (ROI) backtracking, and dual-modality tracing. 
  Concretely, we employed background modeling to filter the motion information and left the static information for later vehicle detection. 
  For the vehicle detection and tracking module, we adopted YOLOv5 and multi-scale tracking to localize the anomalies. 
  Besides, we utilized the frame difference and tracking results to identify the road and obtain the mask. 
  In addition, we introduced multiple similarity estimation metrics to refine the anomaly period via backtracking. 
  Finally, we proposed a dual-modality bilateral tracing module to refine the time further. 
  The experiments conducted on the Track 4 testset of the NVIDIA 2021 AI City Challenge yielded a result of 0.9302 F1-Score and 3.4039 root mean square error (RMSE), indicating the effectiveness of our framework.
\end{abstract}
\vspace{0.35cm}

  \end{@twocolumnfalse} 
] 



\section{Introduction}
\let\thefootnote\relax\footnote{* These authors contributed equally to this work.}
{\let\thefootnote\relax\footnote{{$\dag$Corresponding author: Zhenzhong Chen, (E-mail:  \texttt{zzchen@ieee.org}). This work was supported in part by National Key R\&D Program of China under contract No. 2018YFB0505500 and 2018YFB0505501, and the Fundamental Research Funds for the Central Universities (Grant No. 2042020kf0205).}}}

Traffic anomaly detection, one of the critical components for ITS, attracted more attention as the broader use of surveillance cameras. 
With the aid of anomaly detection, traffic management can respond to emergencies and make instantaneous decisions such as route re-planning and healthcare resource allocation. 
Admitting more and more cameras are mounted within the urban areas, only a few of the data collected by these devices is processed and responded to. 
This is because of the extremely un-even situation between limited human monitoring resources and the vast amount of collected data. 
Thus, an anomaly detection framework of high generalization and efficiency is urgently needed to confront this dilemma. 
\begin{figure}[ht]
	\centering
	 \subfloat{
	   \begin{minipage}[b]{0.48\linewidth}
		 \includegraphics[width=1\textwidth]{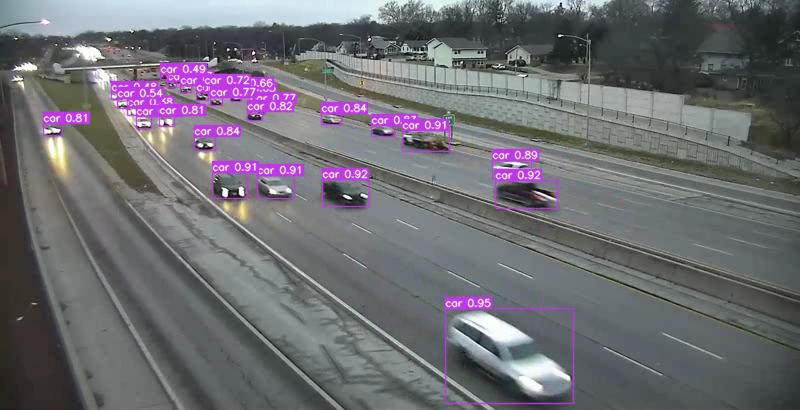} \vspace{6pt}
		 \includegraphics[width=1\textwidth]{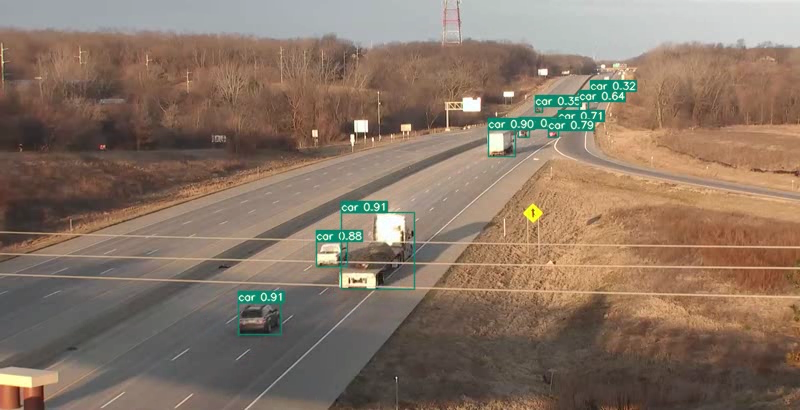}
	   \end{minipage}
	 }
	 \subfloat{
	   \begin{minipage}[b]{0.48\linewidth}
		 \includegraphics[width=1\textwidth]{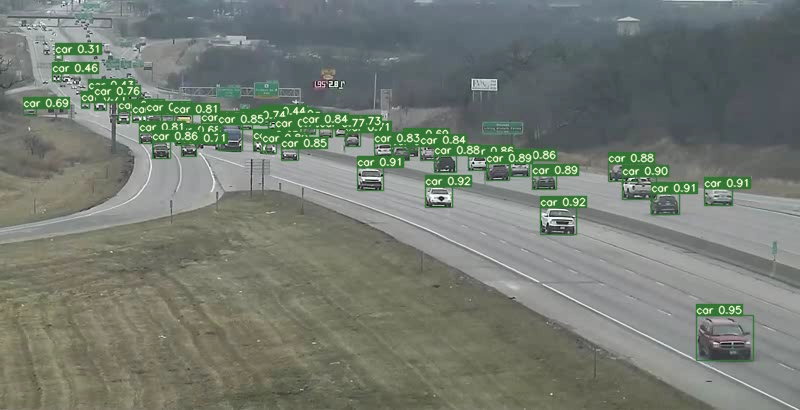} \vspace{6pt}
		 \includegraphics[width=1\textwidth]{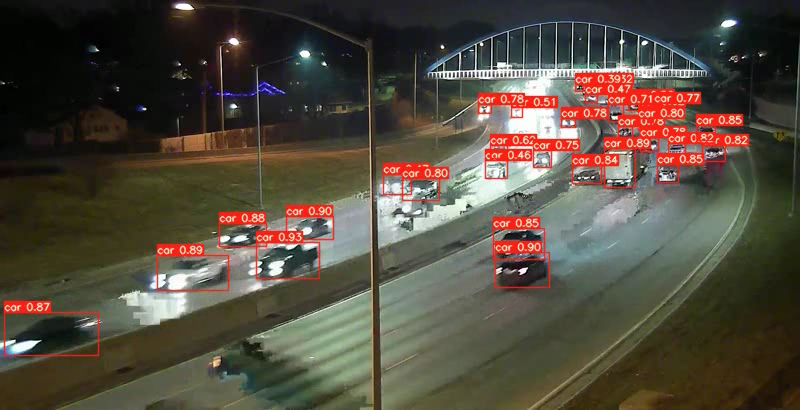}
	   \end{minipage}
	 }
	 \caption{Examples of visualized detection results. Through YOLOv5, even small vehicles in the remote regions can be infered accurately with relatively high confidence score and tight bounding box.
	 }
	\label{yyc_yolo}
\end{figure}
Despite the importance of such an anomaly detection algorithm, several physical challenges, including object occlusion, lighting conditions, ever-changing weather, various anomaly scenarios, etc., must be addressed. 
Another difficulty comes from the distribution of anomaly since it often refers to an event that is not expected to occur in a given context, suggesting the imbalance between normal and abnormal samples. 
This creates great difficulty in precisely identifying the anomaly event in all samples since it only makes up a relatively minor fraction. 
Besides, both the normal scene and abnormal scene are highly diversified. 
The traffic flow pattern for the normal sample can vary from each other, so it is for the abnormal samples. 
Therefore, it is critical to define the traffic anomaly.
Specifically, in the NVIDIA 2021 AI City Challenge, the traffic anomaly can be substantially categorized into two kinds: stalled vehicles and crashing ones.

In recent years, most of the work is deep-learning-based, and much state-of-art work is end-to-end algorithms. 
Such algorithms aim to detect anomalies by modeling normal and abnormal samples through deep neuron networks. 
Admitting great performance has been achieved by these previous work on many datasets, such as UCF-Crime~\cite{UCF-Crime}, ShanghaiTech~\cite{liu2018ano_pred}, UCSD Ped~\cite{10.1145/2683483.2683556}, SUBWAY~\cite{SUBWAY}, they can hardly generalize to another dataset.
For one aspect, the scenes within the aforementioned dataset are often identical and homogenous. 
For another aspect, the anomalies in these datasets give great change in both intensity and motion of the frames. 
Both these two features make the modeling of the anomaly event readily. 
Contrary to these datasets, Iowa DOT~\cite{Naphade18AIC18}, which functions as the benchmark for the challenge, consists of multi-scenarios and various weather, indicating the infeasibility of implementing aforementioned methodologies. 
Facing the current challenges, we proposed a dual-modality framework to fully utilize the motion information among frames to identify the traffic anomaly. 
Generally, two reasonable assumptions to tackle anomaly localization are made. 
First, the normal vehicles should drive along the road without parking movement lasting longer than the traffic signal control period or un-smoothed trajectory; thus, we define the problem as identify the stationary vehicle and vehicles with abnormal trajectory, which often refers to a car crashing event. 
Secondly, either stalled and crashed vehicles will ultimately come to a stop. 

With these two assumptions, we will localize the stopping time of the vehicle then construct the trajectory to determine if it is a stalled vehicle. 
For a stalled vehicle with an un-smoothed trajectory, we determined that the start time would be when it began to deviates from the normal status. 
Concretely, we firstly utilized the classic MOG2~\cite{zivkovic2004improved} algorithm for the background modeling to leave out the static information within the context.
After the background modeling module, we were able to better cope with the instant occlusion of the abnormal vehicles, which managed to make the detection result of these vehicles given by YOLOv5~\cite{glenn_jocher_2021_4418161} more robust for multiple scenarios. 
Then we introduced a novel box-level branch and fused it with a pixel-level branch inspired by~\cite{Bai_2019_CVPR_Workshops, Li_2020_CVPR_Workshops} for the comprehensiveness of vehicle tracking. 
Besides, the mask construction based on the fusion of motion and tracking trajectory was employed to filter out the parking lots, which do not count as anomalous regions. 
In our methodology, the filtered multi-granularity tracking result only provides a rough time-localization for the traffic anomaly. With the time and location provided by the aforementioned module, we refined the result with a proposed novel mechanism named Region of Interest (ROI) backtracking, which adopts multi-faceted distance metric for both object-level and pixel-level. 
Finally, we proposed another novel module, a dual-modality bilateral trajectory tracing branch, which makes a significant contribution to the accurate inference of anomaly time.

The main contributions of this paper are summarized as follows:
\begin{itemize}
\item A novel fusion methodology of the box-level and pixel-level tracking branch is proposed. Each branch functions as compensation for detecting anomaly events, which is suggested to be efficient and robust.
\item A novel backtracking methodology that utilizes the object-level detection result to propose an ROI for similarity estimation with three distance metrics is proposed. The extensive testing of ROI backtracking indicates its superiority in accuracy and efficiency
\item A novel dual-modality bilateral trajectory tracing methodology that utilized the RGB image and sparse optical flow inspired by~\cite{xu2018dual} in opposite directions to mine the spacial-temporal information between frames is proposed. This largely resolved the challenge created by the crashing vehicle.
\end{itemize}

To our best knowledge, we are the very first to propose the dual-modality bilateral trajectory tracing in the challenge to resolve the challenge brings about by abnormal moving vehicles. 
Attributing to the aforementioned strengths, our algorithm achieved a result with F1-Score of 0.9302 and an RMSE of 3.4039. 
The test result proved its effectiveness by outperforming many other teams.

\begin{figure*}
	\centering
	\includegraphics[width=7in]{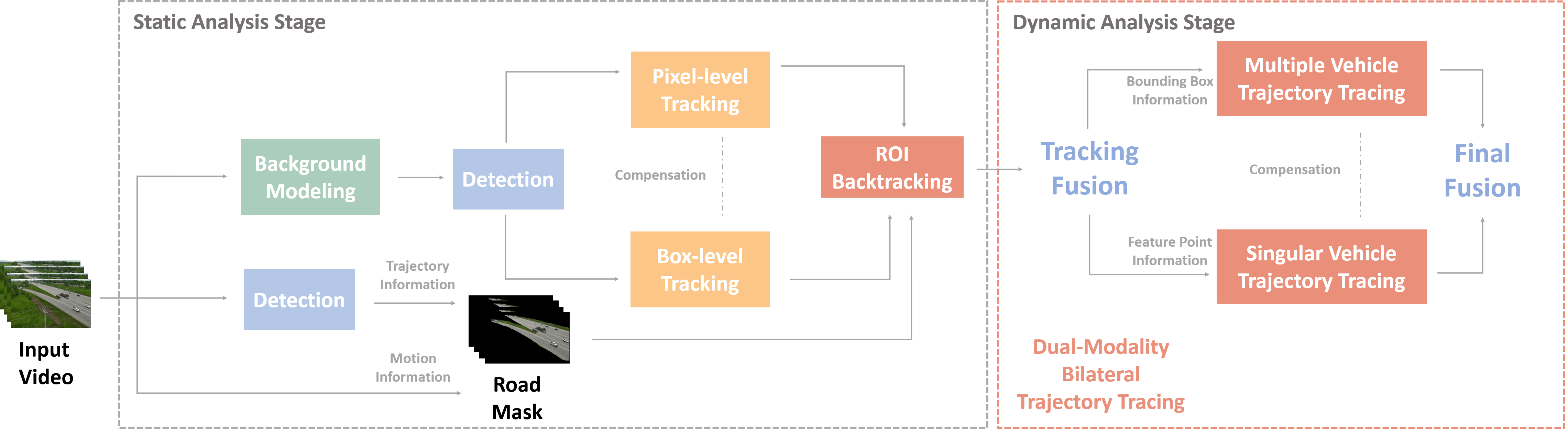}
	\caption{The pipeline of our proposed method that tracking stage provides the exact time and bounding boxes of anomalies in the stationary conditions and localization stage traces back the instants of crashes if possible.}
	\label{yyc_flow_chart}
 \end{figure*}
 

\section{Related Work}\label{sec:related_work}
Anomaly detection, where many researchers have explicitly studied for years, has experienced a significant methodology change.
During the early years, many related research ~\cite{10.1109/ICCV.2011.6126525, 10.1145/2683483.2683556, Saligrama2012VideoAD, wu2010chaotic, cong2011sparse, 6531615, Lu2013AbnormalED, Mahadevan.anomaly.2010, Mehran2009AbnormalCB}focused on how to utilize handcrafted features such as Histogram of Optical Gradient (HOG), Histogram of Optical Flow (HOF), Mixture of Dynamic Textures (MDT)~\cite{Mahadevan.anomaly.2010} that represent both the local and the global pattern of images; then, machine learning methods such as Support Vector Machine (SVM), Markov Random Field (MRF)~\cite{li2009markov, Mahadevan.anomaly.2010}, Isolation Forest~\cite{10.1109/ICDM.2008.17} were implemented to identify the anomaly. 
In the broad sense, these machine learning algorithms mainly face the problem of expensive computation and lack of efficiency.
Deep learning algorithms, as one of the most prospering research fields, have proved their effectiveness on many downstream tasks such as classification, detection, and segmentation, become widely implemented in anomaly detection. Such methods~\cite{UCF-Crime, DBLP:journals/corr/0003CNRD16, hinami2017joint, 10.1145/3123266.3123451, liu2018ano_pred, DBLP:journals/corr/abs-1904-02639, lu2020few, Pang_2020_CVPR, Zhong_2019_CVPR, wu2020not, feng2021mist} often introduce a deep neuron network to learn a good embedding or representation of the normal and abnormal samples. These work can be essentially categorized into two kinds: unsupervised and weakly-supervised framework. 
Unsupervised anomaly detection~\cite{liu2018ano_pred, DBLP:journals/corr/0003CNRD16, hinami2017joint, 10.1145/3123266.3123451, DBLP:journals/corr/abs-1904-02639} only models the normal samples while using multiple loss as the metric to classify the samples since the prediction loss for the abnormal sample is statistically greater than normal ones. The previous work mainly adopted the Auto-Encoder (AE) and Generative Adeversal Network (GAN) as the learning architecture for frame prediction or reconstruction. 
However, these unsupervised work all confront a problem of multiple-possibility for future frame prediction, which results in blurred predictions. Also, to produce robust results, the normal samples must be comprehensive; thus, extensive labeling for samples is inevitable. 
Weakly supervise as another widely-used framework a solution for laborious labeling by using Multiple Instance Learning (MIL) and noised label learning. 
Other deep learning based work includes methodology including multi-modality~\cite{wu2020not}, meta-learning~\cite{lu2020few} also produced considerable performance but still remain as challenging topics. In conclusion, challenges of laborious labeling and lack of interpretability, generalization is still unresolved.
Since many traffic anomalies in Iowa DOT~\cite{Naphade18AIC18} only have subtle intensity change between frames, deep learning method might not be sensitive enough for these anomalies. 
Therefore, in previous AI City Challenges, hybrid frameworks with deep learning techniques were widely employed by teams. 
In 2018 AI City Challenge, Xu \etal~\cite{xu2018dual} firstly introduced background modeling into the challenge and proposed dual-mode modeling to identify the static and mobile anomalies. 
For 2019 AI City Challenge, Bai \etal~\cite{Bai_2019_CVPR_Workshops} developed an extra road mask module and proposed the spatial-temporal information matrix to discriminate the anomalies. 
In 2020 AI City Challenge, Li \etal~\cite{Li_2020_CVPR_Workshops} proposed a multi-granularity tracking module to facilitate static vehicle tracking.  
In this paper, we proposed a novel dual-modality bilateral trajectory tracing mechanism that utilizes both frame intensity and sparse optical flow to trace the trajectory in forward and backward direction, respectively. This mechanism considerably contributes to the fine-grained localization for anomalies with a relatively small RSME of 3.4039 among all results of past teams.

\section{Methodology}\label{sec:model_arch}
The proposed two-stage framework is shown in Figure~\ref{yyc_flow_chart} with its main modules. 
According to the figure, the static analysis stage provides the time and location when a vehicle came to a complete stop. 
In the dynamic analysis stage, the dual-modality bilateral module will be introduced to retrieve the crashing instant for moving anomalies.
In the following sections, we introduce each module in the sequence of processing.

\subsection{Background modeling}
\label{Section_bg_model}
Since both two types of anomalies, including stalled vehicles and crashed vehicles, generally lead to stopped vehicles left on the road after a period of time, the static object analysis in modeling background is widely implemented in anomaly detection. Meanwhile, the fixed perspective in the videos with seldom rotation provides conveniences to model the background in a proper way.

To model the background, we adopted MOG2~\cite{zivkovic2004improved} based on Gaussian Mixture Model (GMM). According to~\cite{Li_2020_CVPR_Workshops} and our experiments, it was more resilient to scene changes and camera shaking than simple moving average methodology~\cite{xu2018dual}. We followed the empirical parameter $T$ set by ~\cite{Li_2020_CVPR_Workshops} where the components of GMM will be updated in the interval of 120 frames at 30 fps.
Besides, it is worth mentioning that since the inner property of MOG2, the time when anomalies appear in the background modeling videos is often postponed compared to the actual timestamp. Hence multiple methodologies for start time tracing were proposed and introduced in the following sections.


\subsection{Detection model}
\label{Section_detect_model}
With the development of computer vision, object detection is of ever-growing importance. ITS raises many crucial topics such as speed estimation, vehicle re-identification, and anomaly vehicle detection. To effectively aid ITS, a detection model with high performance is the prerequisite to all aforementioned applications. In general, the detector can be divide into two groups: single-stage and second-stage detector. For a single-stage detector, the head structure of the detector is responsible for prediction of location and classification of the bounding box. Dissimilarly, the classification and prediction task were conducted alternatively.
Since our model implements a tracking by detection framework, the selection of the detection model will primarily affect the performance for the later module. Due to the extra vast amount of labeling data required by classification, we determined YOLOv5~\cite{glenn_jocher_2021_4418161} to be our detector in terms of a trade-off between timely-cost-efficiency and performance. YOLOv5, the latest version among the YOLO series, has achieved considerable performance on COCO~\cite{lin2014microsoft} and other datasets. Also, the Cross-Stage-Partial-Connections (CSP) module proposed to facilitate small object detection in YOLOv4~\cite{bochkovskiy2020yolov4} is suitable since the presence of tiny objects in the dataset.

To identify the vehicles of multi-view and multi-scale, we employed several techniques in training. Firstly, we manually labeled the vehicle class for a fraction of the training set frames. Unlike COCO, we only labeled one class, car, for the anomaly detection as the extensive experiment indicates an increase in mean average precision. Secondly, we adopted the adaptive-sized anchor-box, which is deducted through the statistic distribution of all bounding boxes using the K-Means clustering algorithm. Thirdly, a multi-facet data-augmentation technique including shearing, flipping, mosaic transformation, perspective transformation, etc., was employed for generating final training data. Among all, the mosaic data augmentation proposed in Bochkovskiy~\etal YOLOv4~\cite{bochkovskiy2020yolov4} increase the mean average precision by providing integration of four different contexts.
Finally, we used the weight pre-trained on the COCO dataset as initialization and then finetuned on our dataset. For performance evaluation, the detection model yields a result of 0.90 mAP@0.5 and an 0.87 F1-score at the confidence threshold of 0.457. As a comparison, we trained another detection model based on Faster R-CNN~\cite{ren2016faster} with the same training data; it only achieved 0.80 mAP@0.5. In addition, the inference speed of YOLOv5 reached up to 65 fps without Test-Time-Augmentation (TTA) module, suggesting the superiority in inference efficiency and precision. Figure 2 shows some detection results by YOLOv5.

\subsection{Road mask construction}
\label{Section_road_mask}
Based on the fact that the stationary parking vehicles at the parking lot do not count as anomalies, our algorithm only outputs the static vehicle on/beside the main road as an anomaly to avoid the false alarm. Inspired by~\cite{Li_2020_CVPR_Workshops}, we constructed the road mask in terms of both motion and trajectory information. In detail, the mask region was composed of areas where vehicle tracking results are found, and intensity changes were apparent.

\textbf{\textit{Motion-based road mask.}}
We subtracted the consecutive frames and emphasized the areas of difference. Three hyperparameters were set to handle the changes: $diff$ was used as the subtract result in frame level, the parameter $k$ refers to the interval of implement subtract,  $T1$ refers to the upper bound of changing to avoid unexpected camera shaking and rotation, $T2$ refers to the minimum area of ROI to cope with abnormal local changes. Above all, we accumulated all areas met the conditions to construct the motion-based road mask.

\textbf{\textit{Trajectory-based road mask.}}
To further refine the mask of road, we adopted the DeepSORT~\cite{wojke2017simple} as the multi-object tracking algorithm to track all vehicles detected by the former detection module, where region without any tracking results should not be considered as the regular route and, thus, not an anomalous region.  
For the final fusion, we take the intersection of the masks introduced above to eliminate the false alarms. A for recovery of the original shape of roads, morphological operations including dilation and erosion were implemented as the post-processing module for the mask construction. 

\subsection{Vehicle tracking}
\label{Section_vehicle_tracking}
After the background modeling and detection module, detection results for every frame are collected, but no anomaly inference was yet made. To ensure the comprehensiveness of the anomaly inference, two branches of the tracking algorithm were developed.

\textbf{\textit{Pixel-level tracking.}}
For tracking at pixel-level, we followed the framework proposed by~\cite{Bai_2019_CVPR_Workshops} to analyze anomaly using the spatial-temporal matrix, which represents the dynamic information of stationary vehicles. In this method, spatial-temporal matrices $V_{detected}$, $V_{undetected}$, $V_{state}$, $V_{score}$, $V_{start}$ and $V_{end}$ are calculated iteratively to identify the suspicious anomaly region. Concretely, if the state of one normality region transferred to a suspicious one, pixel-level tracking would compare the region with bounding boxes that were previously given to determine whether an anomaly had happened. 

\textbf{\textit{Box-level tracking.}}
With full adoption to anomaly results provided by the pixel-level tracking branch, we first implement the DeepSORT~\cite{wojke2017simple} as the multi-object tracking algorithm on the samples that are predicted as normal ones, which serve as a supplement for anomaly detection. We set multiple criteria to identify the actual anomaly inferences during this phase while suppressing the false positive ones. Specifically, considering the unavoidable loss of tracked id caused by physical environment, we implemented IoU values as criteria to retrieve the original id. Then, we proposed three criteria: interval, frequency, and stability of each id to leave out the authentic anomaly vehicles.

\begin{figure*}[ht]
	\centering
	 \subfloat{
	   \begin{minipage}[b]{0.3\linewidth}
		 \includegraphics[width=1\textwidth]{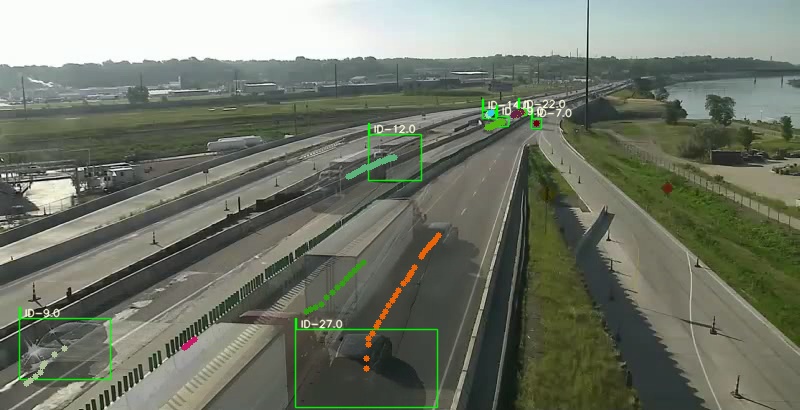} \vspace{3pt}
		 \includegraphics[width=1\textwidth]{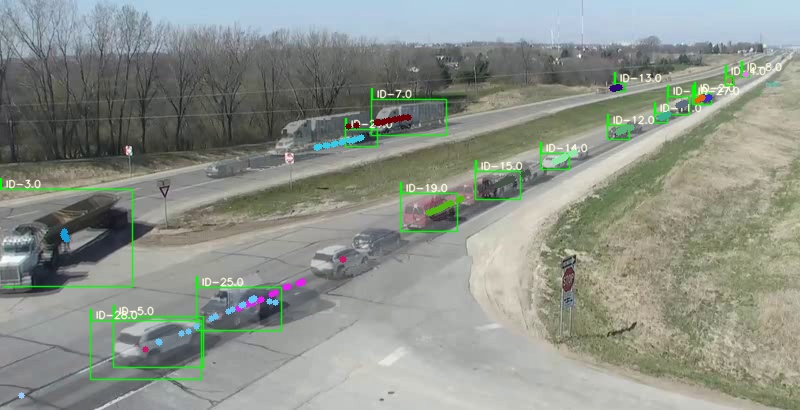}
	   \end{minipage}
	 }
	 \subfloat{
	   \begin{minipage}[b]{0.3\linewidth}
		 \includegraphics[width=1\textwidth]{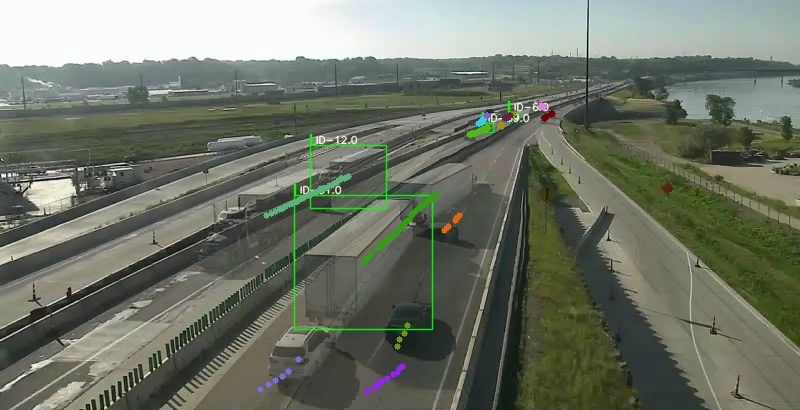} \vspace{3pt}
		 \includegraphics[width=1\textwidth]{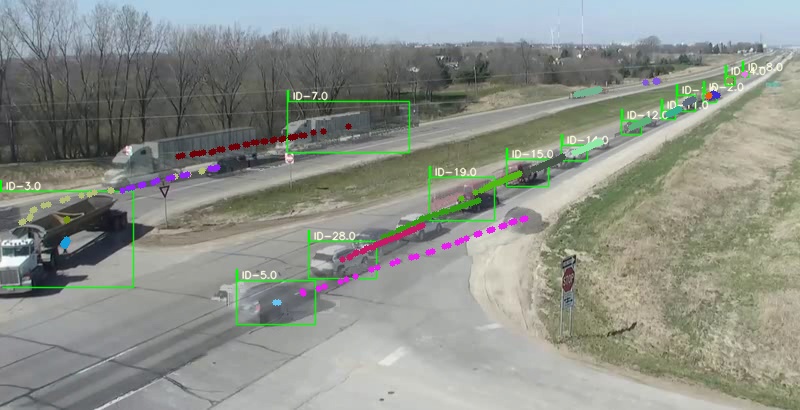}
	   \end{minipage}
	 }
	 \subfloat{
	   \begin{minipage}[b]{0.3\linewidth}
		 \includegraphics[width=1\textwidth]{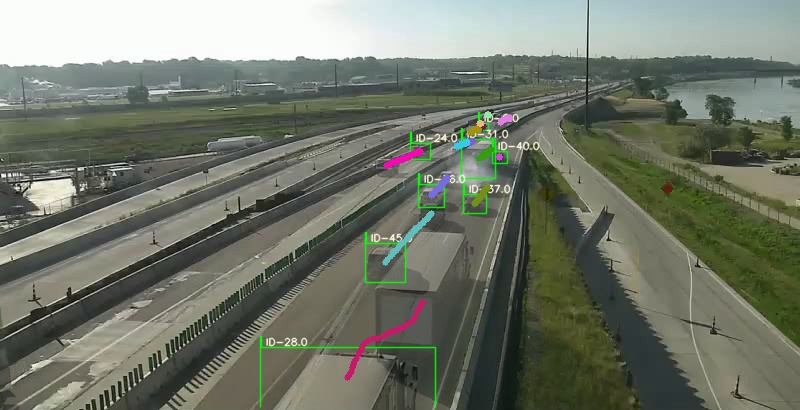} \vspace{3pt}
		 \includegraphics[width=1\textwidth]{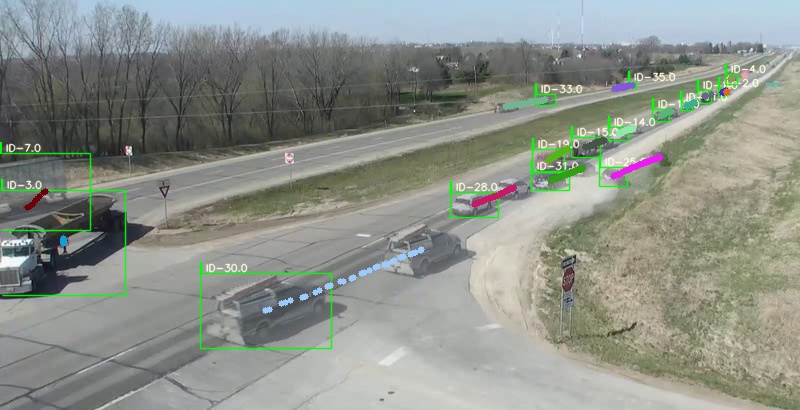}
	   \end{minipage}
	 }
	 \caption{Examples of sharp turning curves. From left to right, trajectories within intervals are displayed in time sequence.
	 }
	\label{yyc_curve}
\end{figure*}

\subsection{ROI backtracking}
\label{Section_ROI_back}
Since the presence of static vehicle appearance delay in background modeling frames, the anomaly inference timestamps for both pixel and box level are also delayed. Therefore, an effective module must be developed to refine the start time of the anomaly.
Despite the imprecise inference time, the location of the anomaly is invariant. Therefore, we can establish a connection between original and background modeling frames by setting an ROI used to estimate the similarity between the same region at two frames over a time span. A novel backtracking methodology was then proposed with the inspiration of~\cite{Li_2020_CVPR_Workshops}. 

Specifically, the ROI was set to be the anomaly bounding box location provided by the previous modules. We firstly search all detection results for each frame on original frames to find the ideal objects, while the IoU value between two bounding boxes functions as the object-level metric.

A pixel-level similarity measurement module then processes the pair of boxes that have high IoU values. A single distance metric might not be sufficient for various scenarios due to its unilaterality; hence, three distance metrics are introduced to evaluate the similarity of the intersection region in adjacent frames, including Peak Signal to Noise Ratio (PSNR), Structural Similarity (SSIM), and Euclidean distance. As the fusion of results produced by metrics above, we utilize a voting and weighting mechanism to reach the best performance while alleviating the extreme deviation caused by the discrepancy among these metrics.

For the anomaly frame contains multiple anomaly vehicles, we filtered the pixel-level inference results and left one with the earliest start time. Then we compared it with the box-level results to select the corresponding nearest results for both branches. After the selection of ROI, we implemented the ROI tracking algorithm separately for results given by two branches. After the ROI tracking module, a simple filtering mechanism is employed to combine similar anomalies and eliminate negative results with time and bounding box information to gain the fused fine-grind tracking results.

\begin{figure*}[t]
	\begin{center}
	\centering
	\includegraphics[width=6.0in]{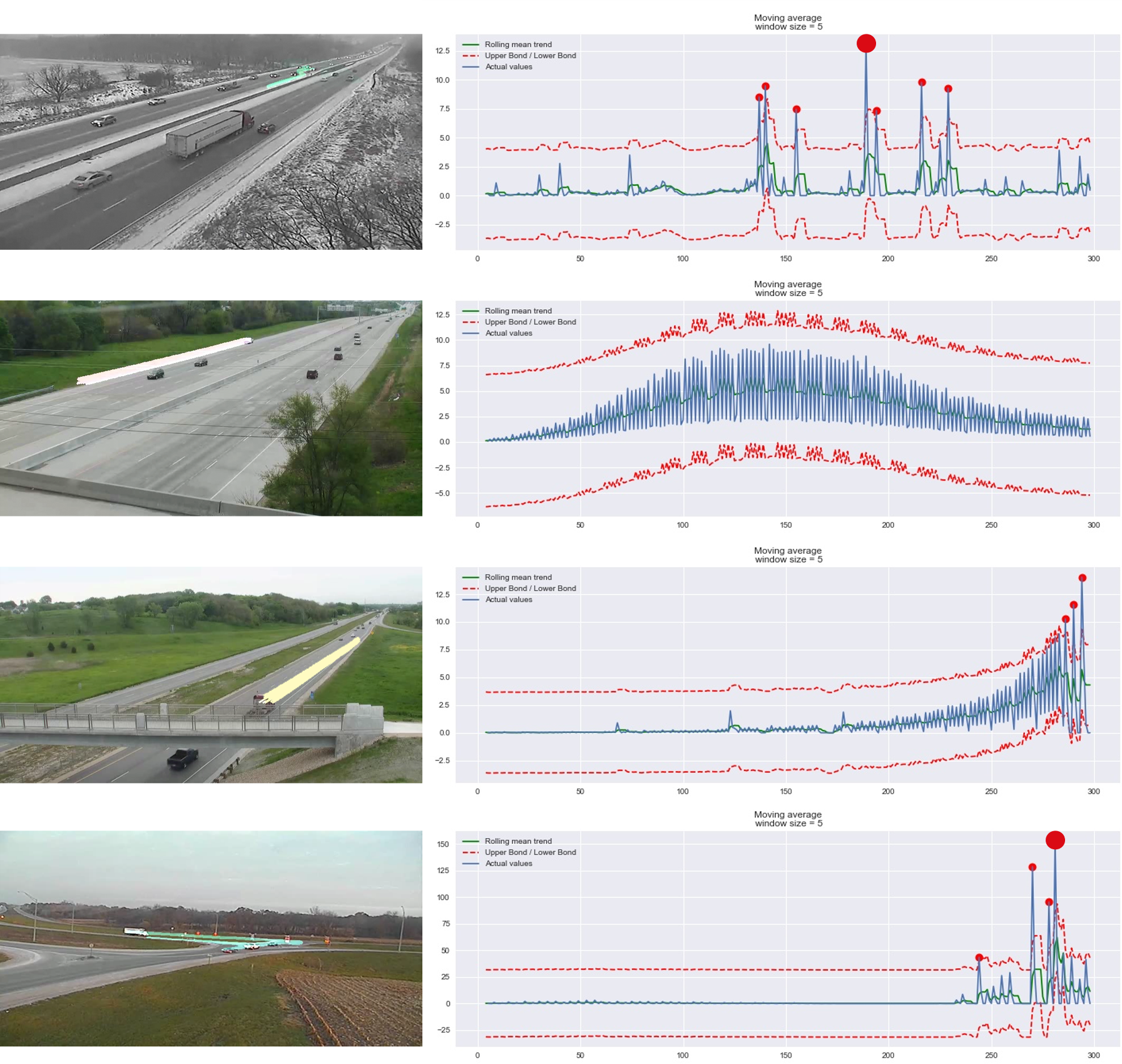}
	\end{center}
	\caption{Examples of singular vehicle trajectory tracing where include stalled and crashed ones. In our hypothesis, once crashes happened, the trajectory of abnormal vehicles will curve immediately which representing violent variation in speed. As the results displayed, speed changes and outliers in the time duration are plotted on the right with optical flow trajectories on the left, respectively.}
	\label{optical_trajectory}
\end{figure*}

\subsection{Dual-modality bilateral trajectory tracing}
\label{Section_fine_grained_localization}
While much previous work succeeded in inferring the accurate start time for stalled vehicles, few have managed inference the crashing instant for moving vehicles since the widely-used background modeling eliminates most of the motion information. Consequently, the aforementioned modules were not adequate for accurate inference of start time for moving vehicles. Therefore, we proposed a novel dual-modality bilateral trajectory tracing module further to analyze the motion pattern of the moving vehicles and aimed to localize the crashing timestamp. Explicitly, intensity information corresponding to the global pattern and sparse optical flow corresponding to the local pattern were utilized in forward and backward directions, respectively. 
\subsubsection{Multiple vehicle trajectory tracing}
Statistically, vehicle crashes often come up with sharp turns, which is the primary reaction of drivers when encountering such anomalies. Also, the sharp turns of an abnormal vehicle often show transitivity in other vehicles along the road for collision avoidance. Therefore, we determined the anomaly time as the interval contains an extensive amount of unsmoothed trajectories.

To obtain the dynamic information, we took the time span $S$, consisting of 26 intervals of 1s as $\{S_0,S_1,...,S_{25}\}$, which starts 20s earlier than the given timestamp and ends at 6s later. The center points of bounding boxes on the original frames are stacked over time to form the trajectory for each vehicle. Each time node $S_i$ corresponds to a set of trajectories for all vehicles in the camera monitoring area, which can be represented as $\{Traj_0^{i},Traj_1^{i},...,Traj_n^{i}\}$.
We consider each normal ${Traj_j}^i$ within 1 second, which is a tolerable interval to be straight.
On the contrary, the abnormal one forms curving trajectories in a short period. Hence, to estimate the curving degree, we utilize the Least Square algorithm to fit straight lines and calculate estimation error ${E_j^i}$ as the curving index. To seek out extreme deviation values in time scale, we set a constant threshold $T$ to calculate abnormal curves in each trajectory ${Traj_j}^i$ where ${E_j^i}$ is higher than $T$ add up for each interval denoted as $N_i$. Finally, a time series of abnormal sharp curves can be established by stacking $N_i$ over time.

Obviously, distractions exist in some scenarios that blending roads contribute to the normal curves of trajectories which should not be misclassified into abnormal ones. In this situation, time series will have an initial value of $N_i$ as in each interval, amount of vehicles pass through the roads, which form platform effect with relatively equal $N_i$ value and impair the peak of time series. Thus We adopt hierarchical constraints to avoid this kind of false-positive detection. Besides, road masks are introduced to eliminate unexpected trajectories out of the main road area and to seek off-track trajectories with length and intersection constraints. 

From the visualization shown in Figure~\ref{yyc_curve}, some crashes can be included and traced accurately to the exact time, align with our sharp turning hypothesis.

\subsubsection{Singular vehicle trajectory tracing}
Admitting the effectiveness of multiple vehicle trajectory tracing, such object-level algorithm with global pattern analysis will fail in cases such as camera shaking and global similar motion pattern. The former case mainly results in unavoidable unstable or vanished detection, making the constant and steady tracking of same vehicle difficult. The later case mainly fail because of lacking of sufficient contrastive normal samples.

Therefore, we proposed another branch of singular trajectory tracing to obtain the local motion pattern captured by sparse optical flow as the compensation for global perspective anomaly time localization. Optical flow, as a classical feature and modality in computer vision, is of great importance for motion analysis. Unlike the object-level trajectory tracing, it remains feasible when experiencing aforementioned cases. To avoid the expensive computational cost led by pixel and its neighborhood-wised tracking, we limited the tracking feature points for each anomaly detected. Concretely, since we had already obtained a refined stopping time and bounding box location given by the ROI backtracking, we used the sparse optical flow tracking algorithm, Lucas–Kanade method, in backward direction. 
With three assumptions that can be fairly made for optical flow: brightness constancy over time, subtle motion and spatial similarity among neighbourhood, the motion of feature points can be reasonably calculated. 
For a point with location $(x,y)$ and its displacement$(dx,dy)$ along horizontal and vertical directions at time stamp $t$ can formulate the brightness constancy function as follows:
\begin{equation}
    \begin{split}
        I(x,y,t)&=I(x+dx,y+dy,t+dt)
    \end{split}
\end{equation}

Then take the Taylor expansion of $I(x+dx, y+dy, t+dt)$ at $(x,y,t)$:
\begin{equation}
    \begin{split}
        I(x,y,t) \approx I(x,y,t) + I_x \cdot \frac{dx}{dt}  + I_y \cdot \frac{dy}{dt} + I_t
    \end{split}
\end{equation}

Finally, the instance motion can be denoted as $(u,v)$ where $u=\frac{dx}{dt}$ and $v=\frac{dy}{dt}$.
Then we initialized $p$ feature points within the bounding box at stopping frame and track these points backward for 390 frames denoted as a sequence of $\{F_0,F_1,...,F_{390}\}$ where each time stamp corresponds to $F_i$. For $p$ tracking results in $F_i$, including position and tracking status, we denoted them as $\{Pos_0^{i},Pos_1^{i},...,Pos_p^{i}\}$ and $\{St_0^{i},St_1^{i},...,St_p^{i}\}$, respectively. 
For next frame, only the points that are successfully tracked in the previous frame should be updated to $p$. Supposing each point location to be $Pos_n$, where only points with $St_n$ equal to true is further tracked. 
The points without being tracked in the previous frame are deposed.

To avoid a small fraction of unreliable optical flow tracking, we adopted the outlier filtering methodology proposed by~\cite{xu2018dual} using the K-Nearest Neighbour to exclude the highly deviated points. Then we calculate the average $u$ and $v$ for every points and gives the final velocity magnitude as follows:
\begin{equation}
  \begin{split}
      m&=\sqrt{\bar{u}^2+\bar{v}^2}
  \end{split}
\end{equation}
The velocity was then stacked as a time series for each vehicle. To further reduce the noise in velocity time series, a suppression method was proposed to pick out the top four velocity values selected with the same interval from the top, represented with its indexes as $\{m_{max}^0,m_{max}^2,m_{max}^4,m_{max}^6\}$ and the variable length of neighbor $L-index$, where $L$ is the constant number set before the procedure. If all the neighbors $\{m_{i-neighbor},m_{i-neighbor+1},...,,m_{i-1},m_{i+1},\\...,m_{i+neighbor}\}$ are smaller than $m_i$ itself, we set $m_i$to 0 to weaken the disturbance of peak noises.

As for the final result, we utilized a moving window estimation module to identify the anomaly. Specifically, we calculated the mean value $\overline{m}$ for the rolling window. The mean absolute error (MAE) and standard deviation between the values in the rolling window and $\overline{m}$ were also computed as $m\_mae$ and $m\_std$. Then the height of normal interval can be determined as $2(m\_mae+scale*m\_std)$ where the centre value is $\overline{m}$.
All points distributed out of the range are sought. Through further judgment, drastic changes and smooth fluctuations are categorized into anomaly and normalcy. We list out some typical distributions of abnormal velocity points in Figure~\ref{optical_trajectory} that extreme pulses such as the top-left curve and bottom-right one are the considered anomalies at the summit, whereas the left two are normal ones with a subtle variation or smooth acceleration.

\section{Experiments}
\subsection{Track 4 dataset}
Iowa DOT~\cite{Naphade18AIC18}, the Track 4 dataset in NVIDIA 2021 AI CITY Challenge is partitioned into 100 training videos and 150 testing videos which contains videos with an approximate length of 15 minutes and at a frame rate of 30 fps and a resolution of $800 \times 410$ as the benchmark for the challenge. 
Some of them contain anomalies due to stalled or crashed vehicles. 
The main target of the challenge is to identify and localize anomalies in test videos to return the start time of the anomaly and confidence score.
\begin{table}
	\centering
		\caption{Our results on Track 4 testset}
		\vspace{3pt}
	\begin{tabular}{c|c|c}
	 \hline
	F1-Score & RMSE & S4-Score\\
	\hline
	0.9302 & 3.4039 & 0.9197\\
	\hline
	\end{tabular}
	\label{table_result}
\end{table}
\subsection{Implementation details}
\textbf{\textit{Detection model.}} YOLOv5 is implemented as the detector of our model. For the optimizer, we adopted Stochastic Gradient Decent with an initial learning rate of $10^{-2}$. The K value and number of anchors are both set to be 9. The inference augmentation is also used in the testing video, where the TTA module will enlarge the image size by 30\%; thus, the size of the image will be expanded to $832 \times 832$. The reason for this is to recognize the smaller object better.

\textbf{\textit{Road mask construction.}}  We follow~\cite{Li_2020_CVPR_Workshops} to set motion-based and trajectory-based parameters except for the filtering area, which is set as 6000 in the motion-based branch to eliminate the continuous shaking frames. What is more, we reduce the times of dilation in order to polish a thinner outline. 

\textbf{\textit{Pixel-level tracking.}} The minimum abnormal duration and suspicious abnormal duration are 60 seconds and 40 seconds, respectively. Other parameters are initialized corresponding to~\cite{Bai_2019_CVPR_Workshops}.

\textbf{\textit{Box-level tracking.}} For anomaly detection, the minimum abnormal duration is set to be 40s, and the vehicle must appear at least four times out of 5 intervals of 10s. For retrieval of vehicle id, IoU threshold is set to be 0.3. For bounding box stability estimation, the standard deviation of the center point must be below three either for $x$ or $y$ coordinate.

\textbf{\textit{ROI backtracking.}} The IoU between the selected bounding box and background modeling detection results is set to 0.9. The maximum deviation time is 12. Backtracking thresholds of PSNR, SSIM, Euclidean are set as 13, 0.4, 0.7, and average thresholds are 10, 0.3, 0.65 with 15 seconds max backtracking time.

\textbf{\textit{Multiple vehicle trajectory tracing.}} The minimum length of trajectory in each interval is 10 points. The threshold of fitting error is 30. The off-track judgment is combined with an area threshold of 40 pixels and a fitting error of 10 at the lowest frequency of 8 times.

\textbf{\textit{Singular vehicle trajectory tracing.}}
The K value for the KNN filter is set to be 6. The density estimation of the clustering points was filtered by a threshold of 6.6. Finally, the scaling factor used to construct the normal interval in moving window estimation is 2.5.
\begin{figure}[!t]
	\begin{center}
	\centering
	\includegraphics[width=3.2in]{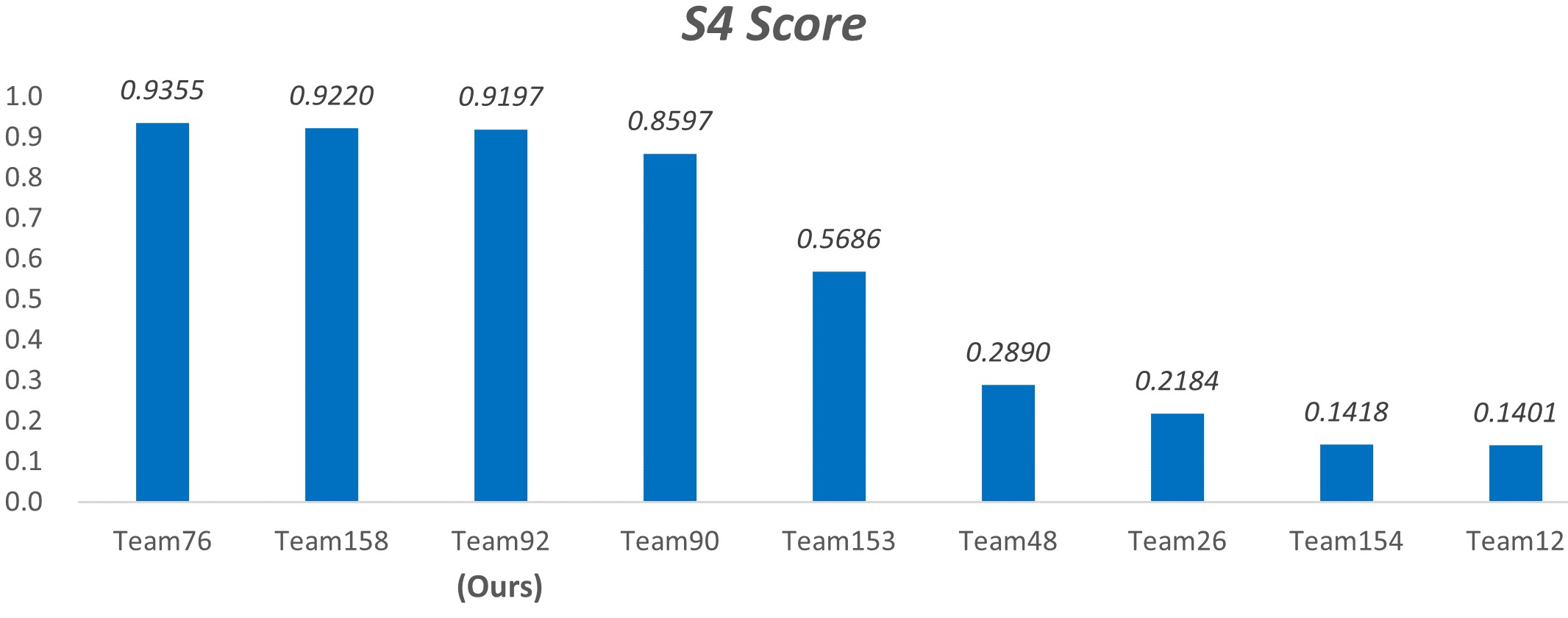}
	\end{center}
	   \caption{The comparison of Track 4 testset results on public leaderboard.}
	\label{rank_image}
\end{figure}

\subsection{Evaluation and experimental results}
Evaluation indexes for algorithm performance consist of $F1$-$Score$ and $NRMSE$ for Track 4, respectively, representing the identification accuracy and time localization error. Specifically, the Track 4 score will be computed as:
\begin{equation}
\begin{split}
    S4&=F1 \times (1-NRMSE)
\end{split}
\label{eval_equation}
\end{equation}
where $F1$ determines the harmonic mean of precision and recall, and $NRMSE$ is the min-max normalization between 0 and 300 frames of time error. $RMSE$ can be computed on difference of time between ground truth and true positive predictions. Briefly speaking, the maximum average prediction error is 300 seconds corresponding to a minimum $NRMSE$ value 0. Specifically, $NRMSE$ can be calculated as:
\begin{equation}
\begin{split}
    NRMSE&=\frac{min(\sqrt{\frac{1}{TP}\sum_{i=1}^{TP} (t_i-{t_i}^{gt})^2},300)}{300}
\end{split}
\end{equation}

We evaluate our methodology on the Track 4 testing data and obtain relatively high $F1$-$Score$ at 0.9302 while maintaining low $RMSE$ value at 3.4039, referring average time error in seconds. Finally, according to Equation~\ref{eval_equation}, we gain 0.9197 $S4$-$Score$ on the full test set. 

\section{Conclusion}
In this paper, a novel dual-modality two-stage anomalies detection model that analysis both static and motion information for abnormal vehicles was proposed. The static analysis stage infers the time and location of the static anomalies at both pixel and box level; As a supplementary, the dynamic analysis stage, including multiple and singular trajectory tracing branch, was proposed to infer the crashing instant in bilateral direction. In NVIDIA 2021 AI City Challenge, a considerable performance with an F1-Score of 0.9302 and RMSE of 3.4039 was achieved through our proposed framework. 



\normalsize
\bibliography{egbib}

\begin{thebibliography}{33}
\providecommand{\natexlab}[1]{#1}
\providecommand{\url}[1]{\texttt{#1}}
\expandafter\ifx\csname urlstyle\endcsname\relax
  \providecommand{\doi}[1]{doi: #1}\else
  \providecommand{\doi}{doi: \begingroup \urlstyle{rm}\Url}\fi

\bibitem[Sultani et~al.(2018)Sultani, Chen, and Shah]{UCF-Crime}
Waqas Sultani, Chen Chen, and Mubarak Shah.
\newblock Real-world anomaly detection in surveillance videos.
\newblock In \emph{Proceedings of the IEEE Conference on Computer Vision and
  Pattern Recognition}, pages 6479--6488, 2018.

\bibitem[Liu et~al.(2018)Liu, Luo, Lian, and Gao]{liu2018ano_pred}
Wen Liu, Weixin Luo, Dongze Lian, and Shenghua Gao.
\newblock Future frame prediction for anomaly detection--a new baseline.
\newblock In \emph{Proceedings of the IEEE Conference on Computer Vision and
  Pattern Recognition}, pages 6536--6545, June 2018.

\bibitem[Biswas and Babu(2014)]{10.1145/2683483.2683556}
Sovan Biswas and R.~Venkatesh Babu.
\newblock Short local trajectory based moving anomaly detection.
\newblock In \emph{Proceedings of the 2014 Indian Conference on Computer Vision
  Graphics and Image Processing}. Association for Computing Machinery, 2014.
\newblock ISBN 9781450330619.
\newblock \doi{10.1145/2683483.2683556}.
\newblock URL \url{https://doi.org/10.1145/2683483.2683556}.

\bibitem[Adam et~al.(2008)Adam, Rivlin, Shimshoni, and Reinitz]{SUBWAY}
Amit Adam, Ehud Rivlin, Ilan Shimshoni, and Daviv Reinitz.
\newblock Robust real-time unusual event detection using multiple
  fixed-location monitors.
\newblock \emph{IEEE Transactions on Pattern Analysis and Machine
  Intelligence}, 30\penalty0 (3):\penalty0 555--560, 2008.
\newblock \doi{10.1109/TPAMI.2007.70825}.

\bibitem[Naphade et~al.(2018)Naphade, Chang, Sharma, Anastasiu, Jagarlamudi,
  Chakraborty, Huang, Wang, Liu, Chellappa, Hwang, and Lyu]{Naphade18AIC18}
Milind Naphade, Ming-Ching Chang, Anuj Sharma, David~C. Anastasiu, Vamsi
  Jagarlamudi, Pranamesh Chakraborty, Tingting Huang, Shuo Wang, Ming-Yu Liu,
  Rama Chellappa, Jenq-Neng Hwang, and Siwei Lyu.
\newblock The 2018 nvidia ai city challenge.
\newblock In \emph{Proceedings of the IEEE Conference on Computer Vision and
  Pattern Recognition Workshops}, pages 53--–60, 2018.

\bibitem[Zivkovic(2004)]{zivkovic2004improved}
Zoran Zivkovic.
\newblock Improved adaptive gaussian mixture model for background subtraction.
\newblock In \emph{Proceedings of the 17th International Conference on Pattern
  Recognition, 2004. ICPR 2004.}, volume~2, pages 28--31. IEEE, 2004.

\bibitem[Jocher et~al.(2021)Jocher, Stoken, Borovec, NanoCode012,
  ChristopherSTAN, Changyu, Laughing, tkianai, yxNONG, Hogan, lorenzomammana,
  AlexWang1900, Chaurasia, Diaconu, Marc, wanghaoyang0106, ml5ah, Doug,
  Durgesh, Ingham, Frederik, Guilhen, Colmagro, Ye, Jacobsolawetz, Poznanski,
  Fang, Kim, Doan, and Yu]{glenn_jocher_2021_4418161}
Glenn Jocher, Alex Stoken, Jirka Borovec, NanoCode012, ChristopherSTAN, Liu
  Changyu, Laughing, tkianai, yxNONG, Adam Hogan, lorenzomammana, AlexWang1900,
  Ayush Chaurasia, Laurentiu Diaconu, Marc, wanghaoyang0106, ml5ah, Doug,
  Durgesh, Francisco Ingham, Frederik, Guilhen, Adrien Colmagro, Hu~Ye,
  Jacobsolawetz, Jake Poznanski, Jiacong Fang, Junghoon Kim, Khiem Doan, and
  Lijun Yu.
\newblock {ultralytics/yolov5: v4.0 - nn.SiLU() activations, Weights \& Biases
  logging, PyTorch Hub integration}, January 2021.
\newblock URL \url{https://doi.org/10.5281/zenodo.4418161}.

\bibitem[Bai et~al.(2019)Bai, He, Lei, Wu, Zhu, Sun, and
  Yan]{Bai_2019_CVPR_Workshops}
Shuai Bai, Zhiqun He, Yu~Lei, Wei Wu, Chengkai Zhu, Ming Sun, and Junjie Yan.
\newblock Traffic anomaly detection via perspective map based on
  spatial-temporal information matrix.
\newblock In \emph{Proceedings of the IEEE Conference on Computer Vision and
  Pattern Recognition Workshops}, June 2019.

\bibitem[Li et~al.(2020)Li, Wu, Bai, Yang, Tan, Li, Wen, Zhang, and
  Ding]{Li_2020_CVPR_Workshops}
Yingying Li, Jie Wu, Xue Bai, Xipeng Yang, Xiao Tan, Guanbin Li, Shilei Wen,
  Hongwu Zhang, and Errui Ding.
\newblock Multi-granularity tracking with modularlized components for
  unsupervised vehicles anomaly detection.
\newblock In \emph{Proceedings of the IEEE Conference on Computer Vision and
  Pattern Recognition Workshops}, June 2020.

\bibitem[Xu et~al.(2018)Xu, Ouyang, Cheng, Yu, Xiong, Ng, Pranata, Shen, and
  Xing]{xu2018dual}
Yan Xu, Xi~Ouyang, Yu~Cheng, Shining Yu, Lin Xiong, Choon-Ching Ng, Sugiri
  Pranata, Shengmei Shen, and Junliang Xing.
\newblock Dual-mode vehicle motion pattern learning for high performance road
  traffic anomaly detection.
\newblock In \emph{Proceedings of the IEEE Conference on Computer Vision and
  Pattern Recognition Workshops}, pages 145--152, 2018.

\bibitem[Antic and Ommer(2011)]{10.1109/ICCV.2011.6126525}
Borislav Antic and Bjorn Ommer.
\newblock Video parsing for abnormality detection.
\newblock In \emph{Proceedings of the International Conference on Computer
  Vision}, page 2415–2422. IEEE Computer Society, 2011.
\newblock ISBN 9781457711015.
\newblock \doi{10.1109/ICCV.2011.6126525}.
\newblock URL \url{https://doi.org/10.1109/ICCV.2011.6126525}.

\bibitem[Saligrama and Chen(2012)]{Saligrama2012VideoAD}
Venkatesh Saligrama and Zhu Chen.
\newblock Video anomaly detection based on local statistical aggregates.
\newblock \emph{Proceedings of the IEEE Conference on Computer Vision and
  Pattern Recognition}, pages 2112--2119, 2012.

\bibitem[Wu et~al.(2010)Wu, Moore, and Shah]{wu2010chaotic}
Shandong Wu, Brian~E Moore, and Mubarak Shah.
\newblock Chaotic invariants of lagrangian particle trajectories for anomaly
  detection in crowded scenes.
\newblock In \emph{Proceedings of the IEEE computer society Conference on
  Computer Vision and Pattern Recognition}, pages 2054--2060. IEEE, 2010.

\bibitem[Cong et~al.(2011)Cong, Yuan, and Liu]{cong2011sparse}
Yang Cong, Junsong Yuan, and Ji~Liu.
\newblock Sparse reconstruction cost for abnormal event detection.
\newblock In \emph{Proceedings of the IEEE Conference on Computer Vision and
  Pattern Recognition}, pages 3449--3456. IEEE, 2011.

\bibitem[Li et~al.(2014)Li, Mahadevan, and Vasconcelos]{6531615}
Weixin Li, Vijay Mahadevan, and Nuno Vasconcelos.
\newblock Anomaly detection and localization in crowded scenes.
\newblock \emph{IEEE Transactions on Pattern Analysis and Machine
  Intelligence}, 36\penalty0 (1):\penalty0 18--32, 2014.
\newblock \doi{10.1109/TPAMI.2013.111}.

\bibitem[Lu et~al.(2013)Lu, Shi, and Jia]{Lu2013AbnormalED}
Cewu Lu, Jianping Shi, and Jiaya Jia.
\newblock Abnormal event detection at 150 fps in matlab.
\newblock \emph{Proceedings of the IEEE International Conference on Computer
  Vision}, pages 2720--2727, 2013.

\bibitem[Mahadevan et~al.(2010)Mahadevan, LI, Bhalodia, and
  Vasconcelos]{Mahadevan.anomaly.2010}
Vijay Mahadevan, Wei-Xin LI, Viral Bhalodia, and Nuno Vasconcelos.
\newblock Anomaly detection in crowded scenes.
\newblock In \emph{Proceedings of IEEE Conference on Computer Vision and
  Pattern Recognition}, pages 1975--1981, 2010.

\bibitem[Mehran et~al.(2009)Mehran, Oyama, and Shah]{Mehran2009AbnormalCB}
Ramin Mehran, Alexis Oyama, and Mubarak Shah.
\newblock Abnormal crowd behavior detection using social force model.
\newblock \emph{Proceedings of the IEEE Conference on Computer Vision and
  Pattern Recognition}, pages 935--942, 2009.

\bibitem[Li(2009)]{li2009markov}
Stan~Z Li.
\newblock \emph{Markov random field modeling in image analysis}.
\newblock Springer Science \& Business Media, 2009.

\bibitem[Liu et~al.(2008)Liu, Ting, and Zhou]{10.1109/ICDM.2008.17}
Fei~Tony Liu, Kai~Ming Ting, and Zhi-Hua Zhou.
\newblock Isolation forest.
\newblock ICDM '08, page 413–422. IEEE Computer Society, 2008.
\newblock ISBN 9780769535029.
\newblock \doi{10.1109/ICDM.2008.17}.
\newblock URL \url{https://doi.org/10.1109/ICDM.2008.17}.

\bibitem[Hasan et~al.(2016)Hasan, Choi, Neumann, Roy-Chowdhury, and
  Davis]{DBLP:journals/corr/0003CNRD16}
Mahmudul Hasan, Jonghyun Choi, Jan Neumann, Amit~K Roy-Chowdhury, and Larry~S
  Davis.
\newblock Learning temporal regularity in video sequences.
\newblock In \emph{Proceedings of the IEEE Conference on Computer Vision and
  Pattern Recognition}, pages 733--742, 2016.

\bibitem[Hinami et~al.(2017)Hinami, Mei, and Satoh]{hinami2017joint}
Ryota Hinami, Tao Mei, and Shin'ichi Satoh.
\newblock Joint detection and recounting of abnormal events by learning deep
  generic knowledge.
\newblock In \emph{Proceedings of the IEEE International Conference on Computer
  Vision}, pages 3619--3627, 2017.

\bibitem[Zhao et~al.(2017)Zhao, Deng, Shen, Liu, Lu, and
  Hua]{10.1145/3123266.3123451}
Yiru Zhao, Bing Deng, Chen Shen, Yao Liu, Hongtao Lu, and Xian-Sheng Hua.
\newblock Spatio-temporal autoencoder for video anomaly detection.
\newblock In \emph{Proceedings of the 25th ACM International Conference on
  Multimedia}, pages 1933--1941, 2017.

\bibitem[Gong et~al.(2019)Gong, Liu, Le, Saha, Mansour, Venkatesh, and
  Hengel]{DBLP:journals/corr/abs-1904-02639}
Dong Gong, Lingqiao Liu, Vuong Le, Budhaditya Saha, Moussa~Reda Mansour, Svetha
  Venkatesh, and Anton van~den Hengel.
\newblock Memorizing normality to detect anomaly: Memory-augmented deep
  autoencoder for unsupervised anomaly detection.
\newblock In \emph{Proceedings of the IEEE International Conference on Computer
  Vision}, pages 1705--1714, 2019.

\bibitem[Lu et~al.(2020)Lu, Yu, Reddy, and Wang]{lu2020few}
Yiwei Lu, Frank Yu, Mahesh Kumar~Krishna Reddy, and Yang Wang.
\newblock Few-shot scene-adaptive anomaly detection.
\newblock In \emph{Proceedings of the European Conference on Computer Vision},
  pages 125--141. Springer, 2020.

\bibitem[Pang et~al.(2020)Pang, Yan, Shen, Hengel, and Bai]{Pang_2020_CVPR}
Guansong Pang, Cheng Yan, Chunhua Shen, Anton van~den Hengel, and Xiao Bai.
\newblock Self-trained deep ordinal regression for end-to-end video anomaly
  detection.
\newblock In \emph{Proceedings of the IEEE Conference on Computer Vision and
  Pattern Recognition}, June 2020.

\bibitem[Zhong et~al.(2019)Zhong, Li, Kong, Liu, Li, and Li]{Zhong_2019_CVPR}
Jia-Xing Zhong, Nannan Li, Weijie Kong, Shan Liu, Thomas~H. Li, and Ge~Li.
\newblock Graph convolutional label noise cleaner: Train a plug-and-play action
  classifier for anomaly detection.
\newblock In \emph{Proceedings of the IEEE Conference on Computer Vision and
  Pattern Recognition}, June 2019.

\bibitem[Wu et~al.(2020)Wu, Liu, Shi, Sun, Shao, Wu, and Yang]{wu2020not}
Peng Wu, Jing Liu, Yujia Shi, Yujia Sun, Fangtao Shao, Zhaoyang Wu, and Zhiwei
  Yang.
\newblock Not only look, but also listen: Learning multimodal violence
  detection under weak supervision.
\newblock In \emph{Proceedings of the European Conference on Computer Vision},
  pages 322--339. Springer, 2020.

\bibitem[Feng et~al.(2021)Feng, Hong, and Zheng]{feng2021mist}
Jia-Chang Feng, Fa-Ting Hong, and Wei-Shi Zheng.
\newblock Mist: Multiple instance self-training framework for video anomaly
  detection.
\newblock \emph{arXiv preprint arXiv:2104.01633}, 2021.

\bibitem[Lin et~al.(2014)Lin, Maire, Belongie, Hays, Perona, Ramanan,
  Doll{\'a}r, and Zitnick]{lin2014microsoft}
Tsung-Yi Lin, Michael Maire, Serge Belongie, James Hays, Pietro Perona, Deva
  Ramanan, Piotr Doll{\'a}r, and C~Lawrence Zitnick.
\newblock Microsoft coco: Common objects in context.
\newblock In \emph{Proceedings of the European Conference on Computer Vision},
  pages 740--755. Springer, 2014.

\bibitem[Bochkovskiy et~al.(2020)Bochkovskiy, Wang, and
  Liao]{bochkovskiy2020yolov4}
Alexey Bochkovskiy, Chien-Yao Wang, and Hong-Yuan~Mark Liao.
\newblock Yolov4: Optimal speed and accuracy of object detection.
\newblock \emph{arXiv preprint arXiv:2004.10934}, 2020.

\bibitem[Ren et~al.(2016)Ren, He, Girshick, and Sun]{ren2016faster}
Shaoqing Ren, Kaiming He, Ross Girshick, and Jian Sun.
\newblock Faster r-cnn: towards real-time object detection with region proposal
  networks.
\newblock \emph{IEEE Transactions on Pattern Analysis and Machine
  Intelligence}, 39\penalty0 (6):\penalty0 1137--1149, 2016.

\bibitem[Wojke et~al.(2017)Wojke, Bewley, and Paulus]{wojke2017simple}
Nicolai Wojke, Alex Bewley, and Dietrich Paulus.
\newblock Simple online and realtime tracking with a deep association metric.
\newblock In \emph{Proceedings of the IEEE International Conference on Image
  Processing}, pages 3645--3649. IEEE, 2017.

\end{thebibliography}


\end{document}